\documentclass{article}
\usepackage{microtype}
\usepackage[]{graphicx}
\usepackage{subcaption}
\usepackage{booktabs} 

\usepackage{caption, copyrightbox}

\usepackage{amsmath,amsfonts}
\usepackage[linesnumbered,ruled,vlined]{algorithm2e}
\SetKwInput{KwInput}{Input}                
\SetKwInput{KwOutput}{Output}              

\usepackage{times}
\usepackage{soul}
\usepackage{url}
\usepackage{mathtools}
\usepackage[hidelinks]{hyperref}
\usepackage[utf8]{inputenc}
\usepackage{multirow}
\usepackage[switch]{lineno}
\usepackage{xcolor}
\usepackage{bbm}
\usepackage{amssymb}

\usepackage{IJCAI}

\usepackage{amsmath}
\usepackage{amssymb}
\usepackage{mathtools}
\usepackage{amsthm}
\usepackage{url}
\usepackage{algpseudocode}
\usepackage[capitalize,noabbrev]{cleveref}

\theoremstyle{plain}

\theoremstyle{definition}

\theoremstyle{remark}

\usepackage[textsize=tiny]{todonotes}

\newcommand{\name}{{\tt TUQCP}}

\usepackage{amsmath,amsfonts,bm}

\def\eqref#1{equation~\ref{#1}}

\def\1{\bm{1}}

\DeclareMathAlphabet{\mathsfit}{\encodingdefault}{\sfdefault}{m}{sl}
\SetMathAlphabet{\mathsfit}{bold}{\encodingdefault}{\sfdefault}{bx}{n}

\title{Uncertainty Quantification for Collaborative Object Detection Under Adversarial Attacks}

\author{
Huiqun Huang$^1$\footnote{Contact Author}\and
Cong Chen$^2$\and
Jean-Philippe Monteuuis$^2$\and
Jonathan Petit$^2$\and
Fei Miao$^{1}$\\
\affiliations
$^1$University of Connecticut, $^2$Qualcomm\\
\emails
\{huiqun.huang, fei.miao\}@uconn.edu\\
\{congchen, jmonteuu, petit\}@qti.qualcomm.com
}
\begin{document}

\maketitle

\begin{abstract}
Collaborative Object Detection (COD) and collaborative perception can integrate data or features from various entities, and improve object detection accuracy compared with individual perception. However, adversarial attacks pose a potential threat to the deep learning COD models, and introduce high output uncertainty. With unknown attack models, it becomes even more challenging to improve COD resiliency and quantify the output uncertainty for highly dynamic perception scenes such as autonomous vehicles. In this study, we propose the \textbf{T}rusted \textbf{U}ncertainty \textbf{Q}uantification in \textbf{C}ollaborative \textbf{P}erception framework (\name{}). \name{} leverages both adversarial training and uncertainty quantification techniques to enhance the adversarial robustness of existing COD models. More specifically, \name{} first adds perturbations to the shared information of randomly selected agents during object detection collaboration by adversarial training. \name{} then alleviates the impacts of adversarial attacks by providing output uncertainty estimation through learning-based module and  uncertainty calibration through conformal prediction. Our framework works for early and intermediate collaboration COD models and single-agent object detection models. We evaluate \name{} on V2X-Sim, a comprehensive collaborative perception dataset for autonomous driving, 
and demonstrate a 80.41\% improvement in object detection accuracy 
compared to the baselines under the same adversarial attacks. \name{} demonstrates the importance of uncertainty quantification to COD under adversarial attacks. 
\end{abstract}

\section{Introduction}\label{sec:introduction}
Long-range or occlusion issues caused by limited sensing capabilities and inadequate individual viewpoints have been limiting the performance of single-agent object detection models~\cite{roldao20223d}. To enhance the perception capability, collaborative object detection (COD) has been proposed to leverage the viewpoints from multiple agents to enhance the object detection accuracy~\cite{li2022v2x,xu2022opencood,cai2022analyzing} and promote robustness via information sharing. In particular, raw-data-level (early collaboration), feature-level (intermediate collaboration), and decision-level (late collaboration) fusions have demonstrated satisfactory performance in COD. 
Meanwhile, safety-critical systems such as autonomous vehicles require the uncertainty information of the computer vision results for decision-making~\cite{he2023robust}. 
Given this, uncertainty quantification (UQ) techniques (e.g., direct modeling~\cite{Su2022uncertainty}) have been proposed to provide output uncertainty estimation and improve the detection accuracy of existing COD models~\cite{perceptionCBF_corl21,su2024collaborative}. As object detection errors increase, the corresponding uncertainty also increases. However, existing studies of COD models equipped with UQ often assume that the shared information among agents for collaboration perception is trustworthy, rendering the COD model ineffective under adversarial attacks.

Prior studies have shown that a maliciously crafted imperceptible perturbation added on the shared information in COD can significantly alter the object detection result~\cite{tu2021adversarial}, and undermine the trustworthiness of the model output. To mitigate the impact of adversarial attacks, classical methods such as adversarial training~\cite{madry2017towards}, anomaly detection~\cite{alheeti2022lidar}, and intrinsic context consistencies checking~\cite{li2020connecting} have been investigated to enhance models' adversarial resilience. However, these methods either fail to generalize to unseen attackers or fail in imperceptible perturbation, and lack of output uncertainty estimation. Previous studies have demonstrated the efficacy of UQ techniques in improving adversarial resilience across various domains, including single-agent image-based adversarial example detection~\cite{smith2018understanding}, semantic segmentation~\cite{maag2023uncertainty}, and image classification~\cite{ye2024uncertainty}. However, their impact on COD models under adversarial attacks remains unexplored. Enhancing the resilience of existing COD models against adversarial attacks while providing a reliable measure of object detection credibility remains a significant challenge.

To address the above challenge, we propose the \textbf{T}rusted \textbf{U}ncertainty \textbf{Q}uantification in \textbf{C}ollaborative \textbf{P}erception framework (\name{}). Motivated by the capability of UQ techniques in enhancing model adversarial resilience in other applications, \name{} integrates adversarial training and UQ techniques. We focus on white-box attacks, where the attackers have access to the complete structure of \name{} and COD models. White-box access is regarded as the strongest attacker model. Therefore, successfully defending against it indicates greater model robustness of our proposed scheme. In this case, attackers can add perturbations to an agent's to-be-shared information, then the receiving agent uses the perturbed results for COD prediction. 
\name{} mitigates the impact of these perturbations by providing uncertainty estimation for the object detection output of COD models through a learning-based UQ component.
We introduce an additional uncertainty loss term to guide the learning process of COD model under attacks, aiming at reducing the estimated uncertainty and improving the object detection accuracy against adversarial attacks. 
To provide more trustworthy uncertainty estimation, \name{} then calibrates the output uncertainty by conformal prediction.  \name{} is compatible with different types of COD models, such as early and intermediate collaboration, as well as single-agent detection models, hence, shows its effectiveness for improving object detection resiliency.

The key contributions of this work are summarized as follows:

\begin{enumerate}
  \item We design \name{}, a trustworthy Uncertainty Quantification (UQ) framework that leverages both learning-based uncertainty prediction and conformal prediction to improve the resilience of existing collaborative object detection (COD) models against adversarial attacks. We introduce an UQ loss term to reduce the estimated object detection uncertainty and improve the object detection accuracy of COD models against adversarial attacks. Note that our technique can be applied to various COD models. The experiment results demonstrate a 80.41\% improvement in object detection accuracy
  compared to the baseline models under attacks. 

  \item Our \name{} is flexible and can be applied to both early or intermediate collaboration COD models, and also to single-agent object detection models. In addition, the experiment results demonstrate that by integrating adversarial training and UQ techniques together, \name{} is effective against strong white-box inference attacks such as Projected Gradient Descent (PGD). 


\end{enumerate}

\section{Related Work}\label{sec:related_work}

\textbf{Adversarial Attacks on collaborative object detection (COD).} Many COD models were proposed, focusing on improving the object detection accuracy
~\cite{wang2020v2vnet,li2021learning,xu2022v2x,xu2022bridging}. However, these models have been found to be susceptible to adversarial attacks. Indeed, attackers carefully craft the shared information to mislead the object detection results. In particular, most adversarial attacks exploit the vulnerabilities of COD methods by targeting the objectness score~\cite{im2022adversarial,chow2020adversarial} (i.e., the probability that a bounding box contains an object), the bounding box location~\cite{zhang2019towards}, or class label~\cite{xie2017adversarial,lu2017adversarial,yin2022adc}.


\textbf{Defense Strategies for COD under Adversarial Attacks.} To mitigate the attacks against COD, adversarial training (i.e., training the model to output correct results even under attacks such as FGSM~\cite{goodfellow2014explaining} and PGD~\cite{madry2017towards}) is considered one of the most effective defenses. 
However, adversarial training assumes specific attacks and cannot generalize well to unseen attacks~\cite{zhu2023improving}. Recent work also proposed anomaly detection models~\cite{alheeti2022lidar,hau2021shadow,sun2020towards,li2023among} and methods of intrinsic context consistencies checking of the input data~\cite{li2020connecting,ma2021detecting,xiao2019advit,xiao2018characterizing} for COD under adversarial attacks. However, these methods failed against imperceptible perturbations. 

\textbf{Uncertainty Quantification (UQ) for COD and UQ for Adversarial Attacks.} 
UQ techniques have proven effective in enhancing output reliability and improving object detection accuracy in COD~\cite{rss19,perceptionCBF_corl21,he2023robust,su2024collaborative}. Among them, Monte-Carlo dropout method~\cite{miller2018dropout} and deep ensembles method~\cite{lyu2020probabilistic,ovadia2019can} are two of the widely used methods. However, both methods require multiple runs of inference making it impractical for real-time critical tasks. The learning-based direct modeling~\cite{Su2022uncertainty,meyer2020learning,feng2021review} is further proposed for UQ in COD, which requires only a single inference pass. Though the above methods improve the robustness of COD models under varying inputs, they typically assume that the input data are trustworthy, making them vulnerable to adversarial attacks. 

Several existing works have leveraged UQ as an effective strategy to counter adversarial attacks in applications such as tabular datasets~\cite{santhosh2022using}, semantic segmentation~\cite{maag2023uncertainty}, image classification~\cite{ye2024uncertainty}, and out-of-distribution detection~\cite{everett2022improving}. 
\cite{smith2018understanding} examined the uncertainty measures of mutual information~\cite{li2017dropout}, predictive entropy~\cite{rawat2017adversarial} and softmax variance~\cite{feinman2017detecting} for image-based adversarial example detection. However, these methods assumed that adversarial examples lie far from the image manifold, making them ineffective in scenarios where adversarial perturbations are imperceptible but still capable of misleading the results. 

To the best of our knowledge, no existing studies explore UQ as a means to enhance the robustness of COD models under adversarial attacks. This is particularly crucial, as integrating data from multiple agents increases the system's vulnerability to adversarial attacks.
In this study, we study the efficacy of learning-based UQ method on improving the adversarial resilience of existing COD models. Given that the learning-based UQ method lacks rigorous UQ as it may easily overfit the training dataset, we further integrate conformal prediction~\cite{shafer2008tutorial} to calibrate the estimated uncertainty. Experiment results show that our \name{} works well on different COD models and on different adversarial attacks during inference stage.


\section{Framework}\label{sec:framework}

\subsection{Problem Setup}\label{subsec:problem_setup}

\textbf{Terminology.} Assuming that there are $N$ perception agents in each scene. Among them, we assume $M \in [0, N-1]$ random agents are \textit{attackers}, which share carefully-crafted malicious information. The other $N-M-1$ agents share their truly-observed information (named \textit{collaborators}). 
The agent who tries to utilize the shared information from collaborators while protecting itself from attackers is called \textit{ego-agent}.

\textbf{Key Assumption.} In this study, we focus on white-box attacks, where the attackers have access to the whole structure of our proposed framework \name{}. The attackers generate perturbations to their to-be-shared information to ego-agent.



\textbf{Problem Formulation.} Suppose we have a training dataset $\mathcal{D}_1 = \{(x_i, y_i)\}_{i=1}^{N_1}$, a validation dataset $\mathcal{D}_2 = \{(x_i, y_i)\}_{i=1}^{N_2}$, and a testing dataset $\mathcal{D}_3 = \{(x_i, y_i)\}_{i=1}^{N_3}$. $N_1$, $N_2$, and $N_3$ are the number of data samples (e.g., point cloud sequence) in each dataset. $x_i$ is the input of the COD model and $y_i$ is the corresponding ground truth location of the bounding box of target objects in scene $i$. We will omit the index $i$ when there is no conflict.
Assuming that there are $H$ target objects in each scene and $K$ vertices in each bounding box of target objects. Each vertex is represented by $J$ variables. The ground truth bounding box of a random target object $h \in \{1, \dots, H\}$ is represented by $\{y_{h, k}\}_{k=1}^{K}$, $y_{h, k} \in \mathbb{R}^{J}$. We assume that every variable of the convex is independent and follows a single-variate Gaussian distribution. Every vertex is independent from the other vertex.

Given an existing COD model $f_{\theta}$ with parameters $\theta$. $f_{\theta}$ takes in the shared information $F$ from $N$ agents and outputs the classification probability $\hat{p}=\{\hat{p}_{ h}\}_{h=1}^{H}$ by a classification head and the detected location $\hat{y}=\{\{\hat{y}_{h, k}\}_{k=1}^{K}\}_{h=1}^{H}$ of the bounding box of each target object by a location head. The shared information $F$ can either be the raw data $x$ in early collaboration based COD models or the encoded features in the intermediate collaboration based COD models. We design the \name{} framework that integrates adversarial training and UQ technique in this COD model to enhance the resilience of COD model against adversarial attacks. 

More specifically, before the COD model receiving information from surrounding agents, each of the $M$ randomly selected attackers generate the minimum perturbation ${\delta}_m \in \{{\delta}_m\}_{m=1}^{M}$ to their original shared information $F_m \in \{F_m\}_{m=1}^{M}$ to ego-agent such that maximize the object detection classification error $\mathcal{L}_{cls}\left(f_{\theta}\left(F_m+{\delta}_m\right), p\right)$ of the COD model,
\begin{equation}\label{eq:adversarial_training}
\centering
  \begin{aligned}
  \left[{\delta}_m\right]= \underset{{\delta}_m}{\texttt{min}} \;\{{\texttt{arg} \, \texttt{max}} \; \mathcal{L}_{cls}\left(f_{\theta}\left(F_m+{\delta}_m\right), p\right)\},
  \end{aligned}
\end{equation}
where $p$ is the corresponding classification ground truth of each target object in the scene.

\name{} then estimates the object detection uncertainty $\hat{\sigma}=\{\{\hat{\sigma}_{h, k}\}_{k=1}^{K}\}_{h=1}^{H}$, $\hat{\sigma}_{h, k} \in \mathbb{R}^{J \times J}$, of each target object in the scene by the UQ module $\mathcal{F}_{\omega}$ with parameters $\omega$. Specifically, $\mathcal{F}_{\omega}$ first estimates the preliminary object detection uncertainty by learning-based UQ component $\mathcal{F}_{\omega}^{1}$ and calibrates this uncertainty by conformal prediction component $\mathcal{F}_{\omega}^{2}$.

During training stage, the objective of our framework is to find the parameters $[\theta, \omega]$ such that minimizing the loss function $\mathcal{L}$ on training data $\mathcal{D}=\{\mathcal{D}_1, \mathcal{D}_2\}$: 
\begin{equation}\label{eq:optimization}
\centering
  \begin{aligned}
  \left[{\theta}, {\omega}\right]= \underset{\theta, {\omega}}{\texttt{arg} \, \texttt{min}} \; \mathcal{L}\left(\theta, \omega | \mathcal{D} \right).
  \end{aligned}
\end{equation}
The loss $\mathcal{L}$ is a weighted combination of the original loss $\mathcal{L}_{COD}(\theta)$ of COD model and the loss $\mathcal{L}_{UQ}(\omega, \theta)$ of UQ module. $\mathcal{L}_{COD}(\theta)$ includes the regression loss $\mathcal{L}_{reg}(\theta)$ and classification loss $\mathcal{L}_{cls}(\theta)$.
\begin{equation}\label{eq:loss_function}
\centering
  \begin{aligned}
   \mathcal{L}=w_1\mathcal{L}_{reg}(\theta) + w_2\mathcal{L}_{cls}(\theta) + w_3\mathcal{L}_{UQ}(\omega, \theta), 
  \end{aligned}
\end{equation}
where $w_1 \in \mathbb{R}$, $w_2 \in \mathbb{R}$, and $w_3 \in \mathbb{R}$ are the weights adjusting the influence of three loss terms, respectively.

\subsection{Solution Overview} \label{subsec:solution_overview}
\begin{figure*}
 \centering
 \includegraphics[width=0.7\linewidth]{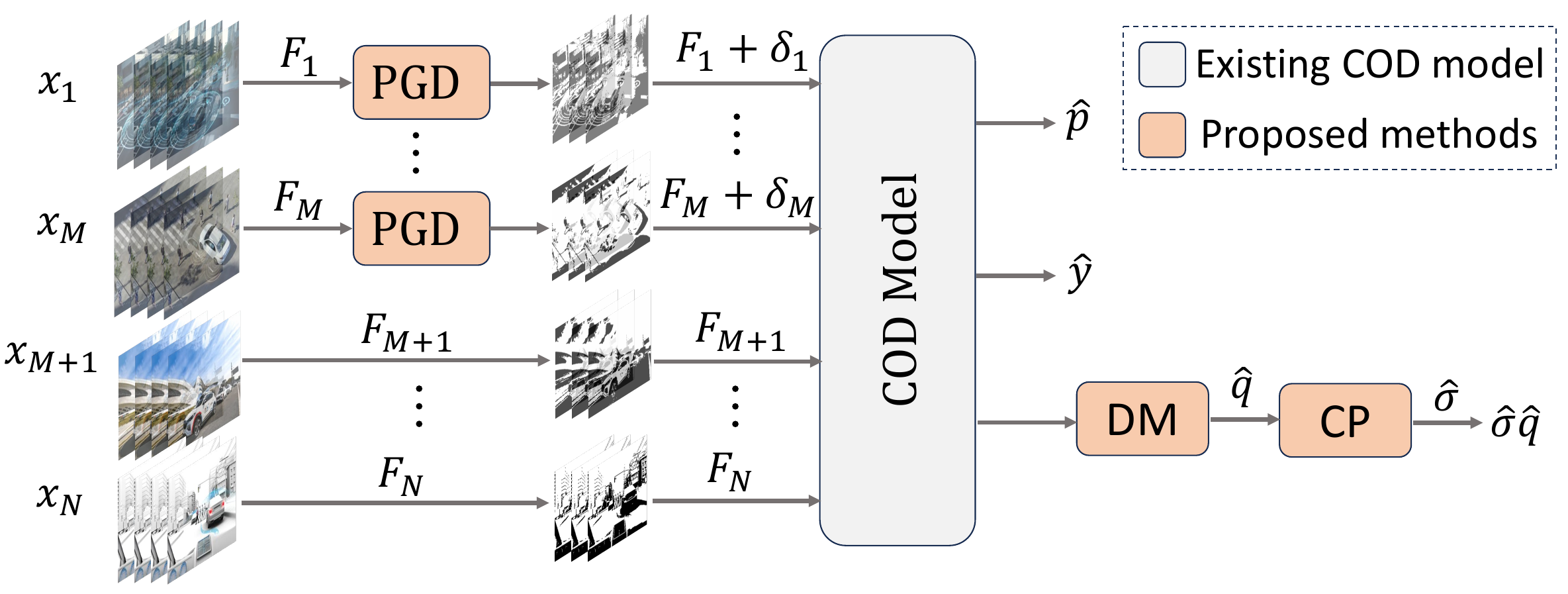} 
  \caption{\textbf{\name{}} randomly selects $M$ agents as attackers and generates minimum perturbation ${\delta}_m$ to the shared information $F_m$ of each attacker to the ego-agent, such thatthe object detection error is maximized. Besides predicting the classification probability $\hat{p}$ and the location $\hat{y}$ of targeted objects by the COD model, \textbf{\name{}} quantifies the preliminary object detection uncertainty $\hat{p}$ of each object through the proposed learning-based UQ (DM) component and calibrates the uncertainty through the proposed conformal prediction (CP) component by the conformal quantile $\hat{q}$.}
  \label{fig:framework}
\end{figure*}

Given an existing COD model, we design the \name{} framework aims at enhancing the resilience of COD model against adversarial attacks. The main structure of \name{} is depicted in Fig.~\ref{fig:framework} and detailed in Alg.~\ref{alg:TUQCP}. \name{} randomly selects $M$ agents among the $N$ collaborative agents as attackers and generates minimum perturbation to the to-be-shared information of each attacker by PGD such that maximize the object detection error. To mitigate the impacts of these perturbations, \name{} quantifies the object detection uncertainty of COD model through the learning-based UQ (DM) component and then calibrates the uncertainty through the conformal prediction (CP) component. During the training stage, \name{} incorporates an additional uncertainty loss term to guide the learning process of the COD model, enhancing its object detection accuracy and reducing the estimated uncertainty under adversarial attacks.

\begin{algorithm}[!ht]

\DontPrintSemicolon

  \KwInput{training dataset $\mathcal{D}_1$ and validation dataset $\mathcal{D}_2$, training iteration $epochs$, input data $F=\{F_n\}_{n=1}^{N}$ from $N$ agents, PGD function, an existing COD model $f_{\theta}$, the proposed learning-based UQ component $\mathcal{F}_{\omega}^{1}$, the proposed conformal prediction component $\mathcal{F}_{\omega}^{2}$, loss function $\mathcal{L}$.}
  \KwOutput{the classification probability $\hat{p}$, the detected location $\hat{y}$, and the calibrated object detection uncertainty $\hat{q}\mathcal{F}_{\omega}^{1}(x)$.}
  
  \For{$epoch=1$ to $epochs$}{
     \For{$(x, y)$ in $\mathcal{D}_1$}{
        Generate perturbations on the shared information of $M$ randomly selected agents by $\{F'\}_{m=1}^{M}=\texttt{PGD}\left(\{F\}_{m=1}^{M}\right)$. Now $F$ includes $\{F'\}_{m=1}^{M}$.
        
        $\hat{p}, \hat{y}=f_{\theta}(F)$
        
        $\mathcal{F}_{\omega}^{1}(F)$
     }

     $\hat{q}=\mathcal{F}_{\omega}^{2}(\mathcal{D}_2, f_{\theta}, \mathcal{F}_{\omega}^{1})$

     Update $\{\theta, \omega\} \leftarrow  \underset{\theta, {\omega}}{\texttt{arg} \, \texttt{min}} \; \mathcal{L}\left(\theta, \omega | \mathcal{D} \right)$
}
\textbf{return} trained $f_{\theta}$, $\mathcal{F}_{\omega}^{1}$, and $\hat{q}$

\caption{Training and validation stages of our \name{} framework.}
\label{alg:TUQCP}
\end{algorithm}
\color{black}


        
        



\subsection{Learning-Based Uncertainty Quantification}\label{subsec:direct_modeling}
In this study, we assume that attackers in COD aim to introduce minimal and imperceptible perturbations into the information they share with the ego-agent, to ultimately mislead the object detection results and undermine output credibility. Methods such as adversarial training, anomaly detection, and intrinsic context consistency checks have been explored to mitigate the effects of adversarial attacks on model performance. However, these methods often struggle to generalize to unseen attacks and lack of effective object detection uncertainty estimation. Recent studies have investigated UQ techniques to provide output credibility for existing COD models. However, these studies often limited to situations where the shared information from surrounding agents are trustworthy. To address this gap, we propose a learning-based UQ component for COD models, which reduces the impact of adversarial perturbations and provides reliable object detection uncertainty estimates. Ideally, higher uncertainty values should correlate with higher misdetection probabilities, offering a more robust measure of detection reliability.

Given an COD model $f_{\theta}$ with parameters $\theta$ and the model inputs $x=\{x_n\}_{n=1}^{N}$ of $N$ agents in a scene, we randomly select $M \in [0, N-1]$ attackers and generate perturbation to their to-be-shared information $\{F_m\}_{m=1}^{M}$ to ego-agent by projected gradient descent (PGD) following Algorithm~\ref{alg:PGD}.

Now having the shared information from $N$ agents, COD model predicts the classification probability $\hat{p}=\{\hat{p}_{ h}\}_{h=1}^{H}$ of $H$ target objects by the classification head and the location $\hat{y}=\{\{\hat{y}_{h, k}\}_{k=1}^{K}\}_{h=1}^{H}$ of the bounding box by the regression head. Our proposed learning-based UQ component $\mathcal{F}_{\omega}^{1}$ quantifies the preliminary object detection uncertainty $\hat{\sigma}=\{\{\hat{\sigma}_{h, k}\}_{k=1}^{K}\}_{h=1}^{H}$ of COD model in parallel with the regression and classification heads. Specifically, $\mathcal{F}_{\omega}^{1}$ shares the same structure as the regression head of COD model. 

To provide trustworthy uncertainty estimation under adversarial attacks, the prediction interval $f_{\theta}(x) \pm \hat{q}\mathcal{F}_{\omega}^{1}(x)$ as introduced in the next section should tend to be narrow while covering the ground truth location of the bounding box. One of the popular approaches to achieve this is reducing the distribution difference between the ground truth and predicted vertices of the bounding box. In this study, we assume the ground truth distribution of each variable of each vertex of the bounding box as a Dirac delta function~\cite{he2019bounding}. We introduce the Kullback–Leibler divergence (KLD)~\cite{meyer2020learning} as additional loss term to guide the learning process of the COD model, reducing the estimated object detection uncertainty and improving the performance of COD model against adversarial attacks. The process is shown as follows,
\begin{align}
    \vspace{-0.1in}
    \label{eq:loss_term_uq}
    \mathcal{L}_{UQ}(\omega, \theta) = \frac{1}{N_1 \times H \times K} {\sum}_{i=1}^{N_1} {\sum}_{h=1}^{H} {\sum}_{k=1}^{K} \\ \left(\frac{(y_{i,h,k}-\hat{y}_{i,h,k})^2}{2\hat{\sigma}_{i,h,k}^2}+log(|\hat{\sigma}_{i,h,k}|)\right).
\end{align}

\begin{algorithm}[!ht]
\DontPrintSemicolon
\caption{Projected Gradient Descent (PGD)}\label{alg:PGD}
\KwInput{Classification loss function $\mathcal{L}_{cls}$ in $\mathcal{L}_{COD}(\theta)$, learning rate $\eta$ of PGD, perturbation budget $\epsilon$, number of iterations $K$, original to-be-shared information $\{F_m\}_{m=1}^{M}$ of $M$ attackers, projection operator $\mathcal{P}_{\mathbb{B}(F,\mathcal{\epsilon})}$, where $\mathbb{B}(F,\mathcal{\epsilon})$ denotes the set of allowed adversarial perturbations, bounded by $\epsilon$.}
 \KwOutput{the adversarial shared information $\{F'_m\}_{m=1}^{M}$.}
Initialization: Initialize the adversarial shared information as $\{F_m\}_{m=1}^{M}$ \\
\For{$k = 0$ to $K-1$}{
    $\mathcal{L}_{cls}(f_{\theta}, \{F'_m\}_{m=1}^{M}, \hat{y})=\mathcal{L}_{cls}(f_{\theta}(\{F'_m\}_{m=1}^{M}), \hat{y})$.
    Update data by gradient descent: $\{F'_m\}_{m=1}^{M} = \{F'_m\}_{m=1}^{M} - \eta \texttt{sign}{\nabla}_{\{F'_m\}_{m=1}^{M}} \mathcal{L}_{cls}(f_{\theta}, \{F'_m\}_{m=1}^{M}, \hat{y})$.
    Project onto feasible set: $\{F'_m\}_{m=1}^{M} = \mathcal{P}_{\mathbb{B}(F,\mathcal{\epsilon})}(\{F'_m\}_{m=1}^{M})$.
}
\textbf{return} $\{F'_m\}_{m=1}^{M}$
\end{algorithm}

\subsection{Conformal Prediction for Uncertainty Calibration}\label{subsec:conformal_prediction}
The learning-based UQ component provides a measurement of the credibility of the object detection result of COD model. However, the learning-based method is prone to overfit the training dataset and gives overconfident uncertainty estimation. To address this, previous studies have explored conformal prediction~\cite{shafer2008tutorial,angelopoulos2023conformal} as a statistical inference method for constructing predictive sets that guarantee a specified probability of covering the true target values. Inspired by this, we propose calibrating the preliminary uncertainty estimation using conformal prediction to enhance the reliability of object detection uncertainty estimation under adversarial attacks.

Conformal prediction generates prediction sets for any model. It holds the explicit, non-asymptotic guarantees that are not contingent upon distributional or model assumptions. In Section~\ref{subsec:direct_modeling}, we assumed that every variable of the vertices of the bounding box of each target object were independent and followed a single-variate Gaussian distribution.  That is, given any input point cloud sequence $x$ and the training and validation data $\mathcal{D}=\{\mathcal{D}_1, \mathcal{D}_2\}$, the corresponding locations follows $Y|x,\mathcal{D}  \sim \mathcal{N}(f_{\theta}(x), \mathcal{F}_{\omega}^{1}(x))$. $\mathcal{F}_{\omega}^{1}(x)$ estimates the preliminary object detection uncertainty of the COD model. 

We train the models $f_{\theta}(x)$ and $\mathcal{F}_{\omega}^{1}(x)$ together to maximize the likelihood of the data. The key idea of conformal prediction is to turn this heuristic uncertainty notion into rigorous prediction intervals by $f_{\theta}(x) \pm \hat{q}\mathcal{F}_{\omega}^{1}(x)$. $\hat{q}$ is the conformal quantile estimated by conformal prediction utilizing the validation dataset.

Given the validation dataset $\mathcal{D}_2 = \{(x_i, y_i)\}_{i=1}^{N_2}$ which has $N_2$ data samples, we estimate the conformal quantile $\hat{q}$ from $\mathcal{D}_2$ by the steps of:
\begin{enumerate}
    \item Design a conformal score function $s(x,y)$ to encode agreement between $x$ and $y$. Smaller scores denote better agreement. Here, we define the score function as:
\begin{equation}\label{eq:conformal_score_function}
\centering
  \begin{aligned}
   s(x,y) = \frac{|y - f_{\theta}(x)|}{\mathcal{F}_{\omega}^{1}(x)}. 
  \end{aligned}
\end{equation}

    \item We compute the conformal quantile $\hat{q}$ as the $\frac{\lceil (N_2+1)(1-\alpha) \rceil}{N_2}$ quantile of the validation scores $S=\{s(x_i,y_i)\}_{i=1}^{N_2}$. $\alpha \in [0,1]$ is the error rate chosen by users. We calibrate the estimated uncertainty from learning-based UQ component by $\hat{q}\mathcal{F}_{\omega}^{1}(x)$. 
    
    \item We then use the estimated conformal quantile $\hat{q}$ to form conformal set $\mathcal{C}(x)=[f_{\theta}(x)-\hat{q}\mathcal{F}_{\omega}^{1}(x), f_{\theta}(x)+\hat{q}\mathcal{F}_{\omega}^{1}(x)$ for any unseen example of testing dataset $\mathcal{D}_3 = \{(x_i, y_i)\}_{i=1}^{N_3}$. Based on split conformal prediction, the main guarantee we can get is that,
\begin{equation}\label{eq:conformal_prediction_guarantee}
\centering
  \begin{aligned}
   \mathbb{P} \left( y \in \mathcal{C}(x) | \mathcal{D}_2\right) \in \left[1-\alpha, 1-\alpha+ \frac{1}{1+N_2}\right). 
  \end{aligned}
\end{equation}
In other words, the probability that the conformal set contains the ground truth location of an object is almost exactly $1-\alpha$.

\end{enumerate}


\section{Experiment}\label{sec:experiment}
\subsection{Experimental Setup} \label{subsec:experimental_setups}

\textbf{Dataset \& Key Setup:} We use the V2X-Sim dataset~\cite{li2022v2x} to verify the efficacy of our \name{}. This dataset contains 80 scenes for training, 10 scenes for validation, and 10 scenes for testing. Each scene of the dataset contains a 20-second traffic flow at a certain intersection with a 5Hz record frequency. This also means that each scene contains 100 time series frames. In addition, 2-5 vehicles are selected to share information to each other in each scene, and 3D point clouds are collected from onboard LiDAR.

In all experiments of this study, the learning rate is set as 0.001, training epochs is 49, learning rate $\eta$ of PGD is set as 0.1, $\epsilon=0.5$, number of iteration $K$ of PGD is 25, number of attackers $M$ in each scene is 2, number of agents $N$ is 6, and error rate of conformal prediction $\alpha$ is 0.1, unless otherwise specified. $\epsilon$ is set as 0.3 for \texttt{When2com}. The host machines are a server with IntelCore i9-10900X processors, four NVIDIA Quadro RTX 6000 GPUs, and one Nvidia A100 with 80 GB RAM. The batch size of the training data when using Nvidia A100 is 20. 

\textbf{Prediction Accuracy Evaluation Metrics:} Following the standard evaluation protocol, we utilize metrics of average precision (AP) at IoU thresholds of 0.5 and 0.7. 

\textbf{Uncertainty Evaluation Metrics:} To verify the performance of \textbf{\name{}} in reducing the estimated uncertainty, we adopt the KLD and
Negative Log-Likelihood (NLL)~\cite{feng2021review} to assess the level of uncertainty in the predicted distribution. Lower values connote a higher degree of precision in uncertainty estimation and narrower uncertainty interval. More specifically, we calculate the average KLD as Eq.~\ref{eq:loss_term_uq} and the average NLL by,
\begin{align}
    \label{eq:nll}
    NLL = \frac{1}{N_3 \times H \times J} {\sum}_{i=1}^{N_1} {\sum}_{h=1}^{H} {\sum}_{j=1}^{J} \\ 
    \left(\frac{(y_{i,h,j}-\hat{y}_{i,h,j})^2}{2\hat{\sigma}_{i,h,j}^2}+\frac{1}{2}log(2\pi{\hat{\sigma}_{i,h,j}}^{2})\right).
\end{align}

\subsection{Baselines} \label{subsec:baselines}
In this study, we employ the following COD models as baselines for comparison. 
\begin{enumerate}
\item \texttt{V2VNet}~\cite{wang2020v2vnet}: \texttt{V2VNet} is an intermediate collaboration based model that propagates agents’ information by a pose-aware graph neural network. It aggregates shared information from other agents by a convolutional gated recurrent unit. After updating features by several rounds of neural message passing, it generates the perception results by the classification and regression heads.
\item \texttt{DiscoNet}~\cite{li2021learning}: \texttt{DiscoNet} is an intermediate collaboration model that uses a directed collaboration graph to highlight the informative spatial regions and reject the noisy regions of the shared information. After updating the features by adaptive message fusion, it outputs the perception results.
\item \texttt{When2com}~\cite{liu2020when2com}: \texttt{When2com} is an intermediate collaboration model that employs attention-based mechanism for communication group construction. It updates the fused feature of all agents by attention-score-based weighted fusion.
\item \texttt{CoMamba}~\cite{li2024comamba}: \texttt{CoMamba} is an intermediate collaboration 3D detection framework designed to leverage state-space models for real-time onboard vehicle perception.   
\item \texttt{Upper-bound}~\cite{li2022v2x}: \texttt{UB}
  is an early collaboration model in which the agents share raw data with each other.
\item \texttt{Lower-bound}~\cite{li2022v2x}: \texttt{LB} is a single-agent perception model which utilizes a single-view perception data for object detection.  
\end{enumerate}

\subsection{Main Results} \label{subsec:main_results} 
\begin{figure*}
\centering
\begin{subfigure}[t]{0.24\textwidth}
    \makebox[0pt][r]{\makebox[7.5pt]{\raisebox{55pt}{\rotatebox[origin=c]{90}{\textbf{DiscoNet(PGD Test)}}}}}%
    \includegraphics[width=\textwidth]{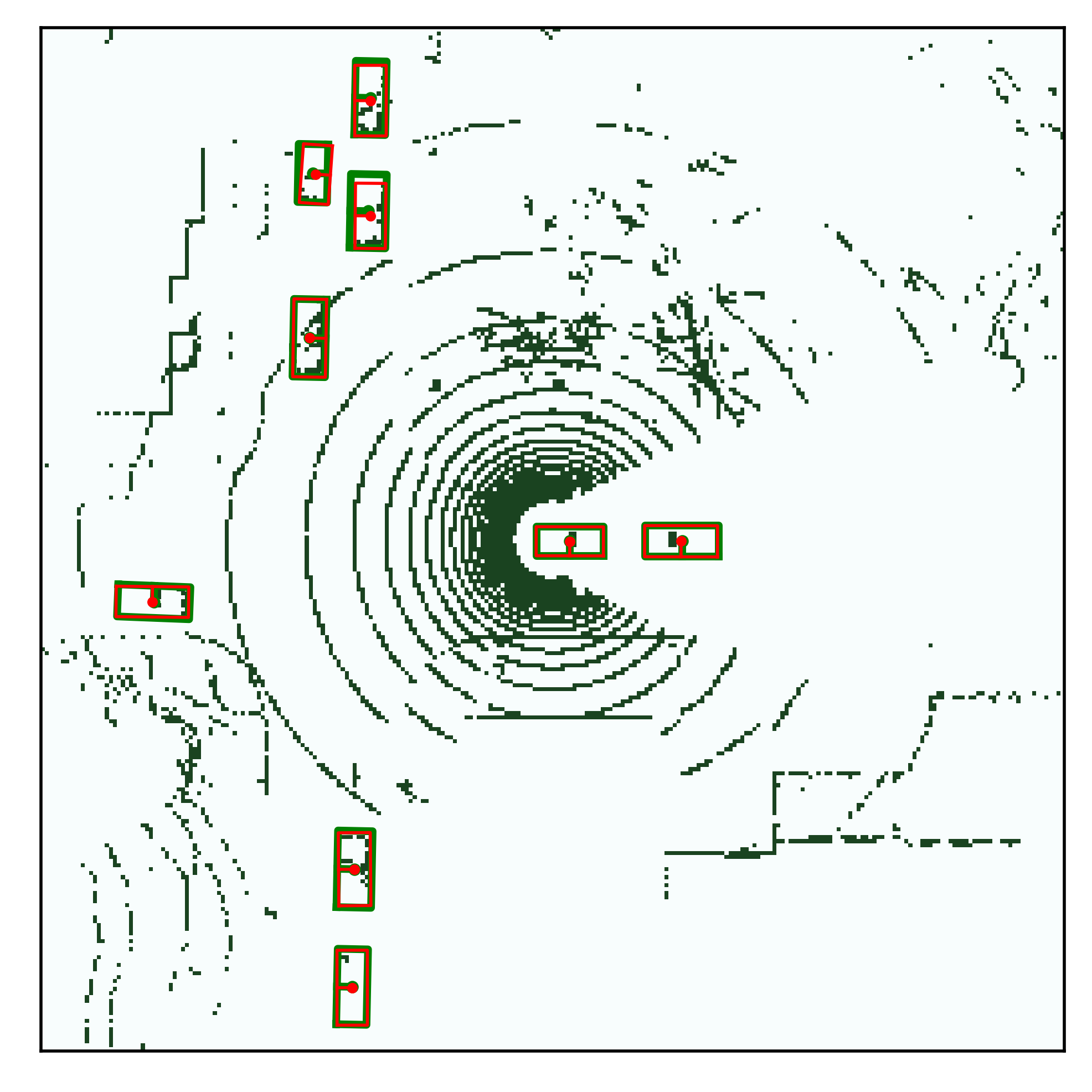}
    \makebox[0pt][r]{\makebox[7.5pt]{\raisebox{55pt}{\rotatebox[origin=c]{90}{\textbf{DiscoNet+TUQCP}}}}}%
    \includegraphics[width=\textwidth]{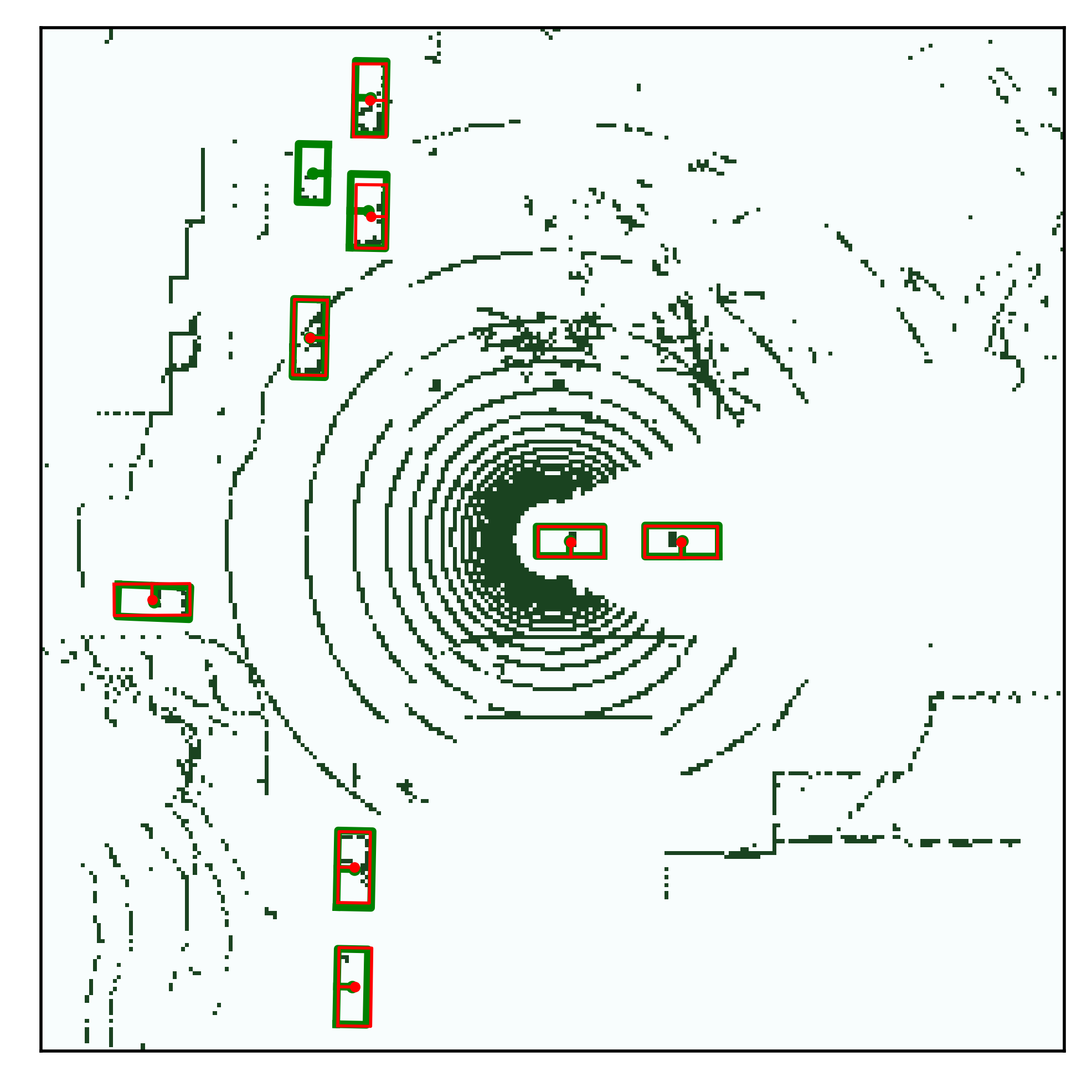}
    \caption{Scene 5 frame 4.}
    \label{fig:scene5_frame4} 
\end{subfigure}
\begin{subfigure}[t]{0.24\textwidth}
    \includegraphics[width=\textwidth]{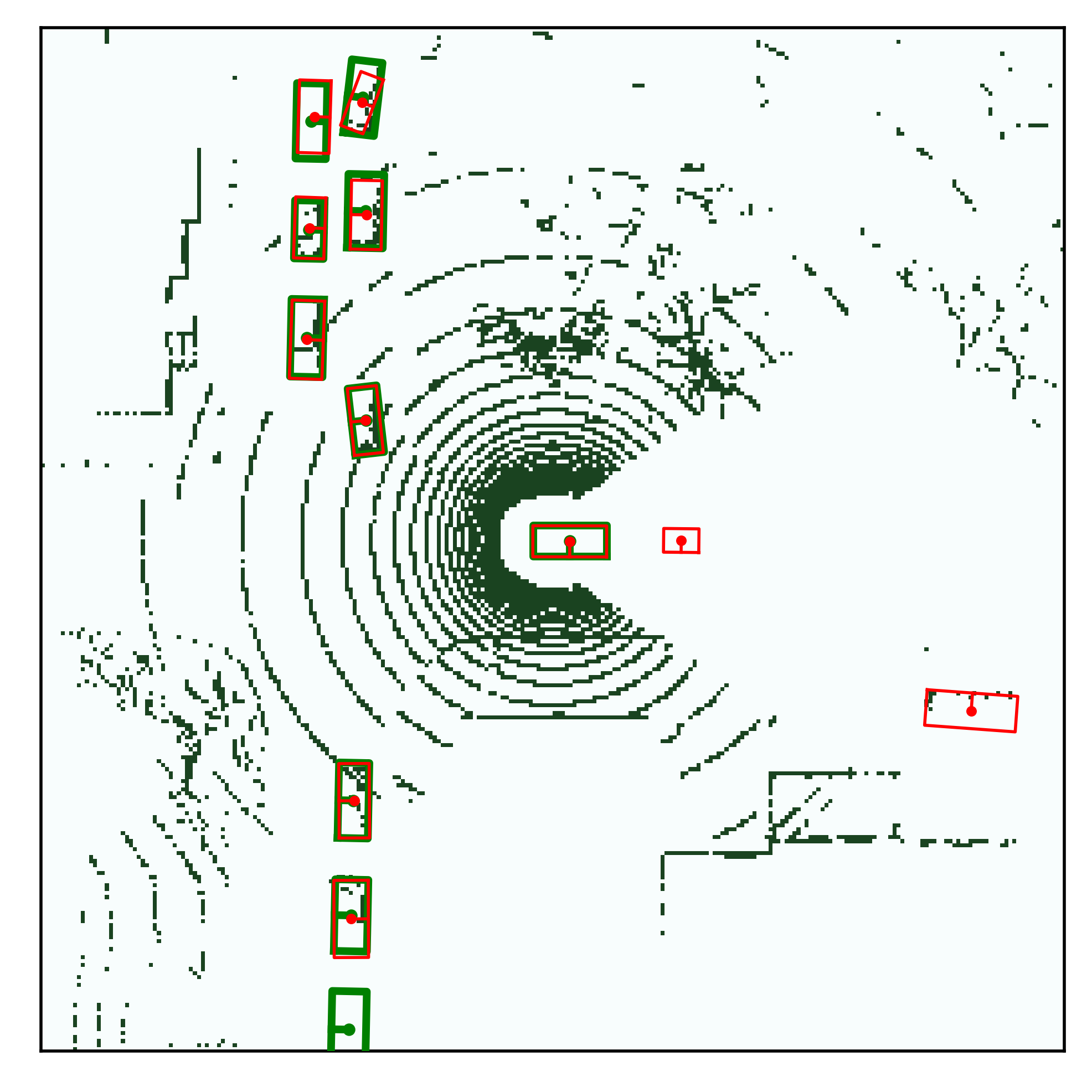}
    \includegraphics[width=\textwidth]{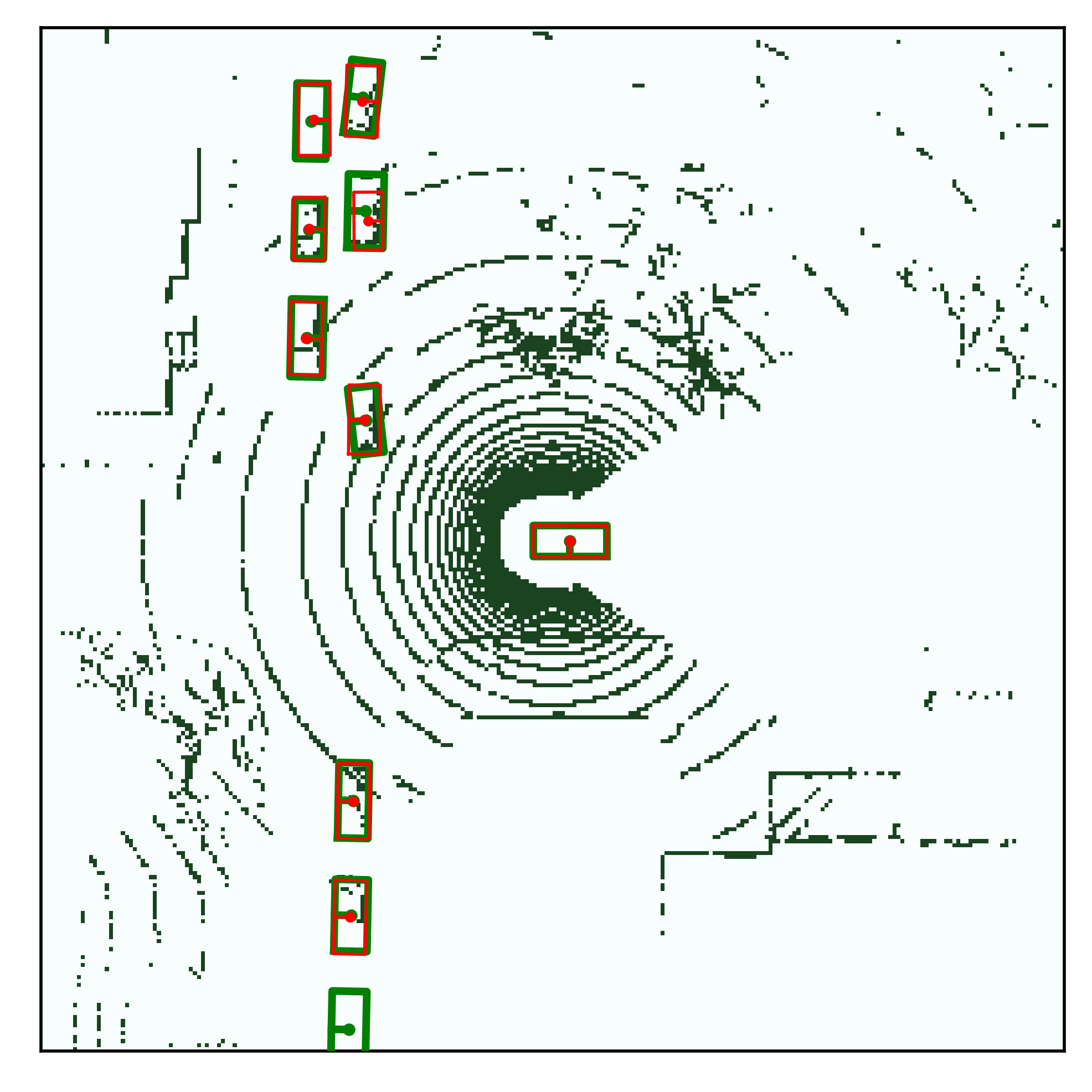}
    \caption{Scene 8 frame 35.}
    \label{fig:scene8_frame35} 
\end{subfigure}
\begin{subfigure}[t]{0.24\textwidth}
    \includegraphics[width=\textwidth]{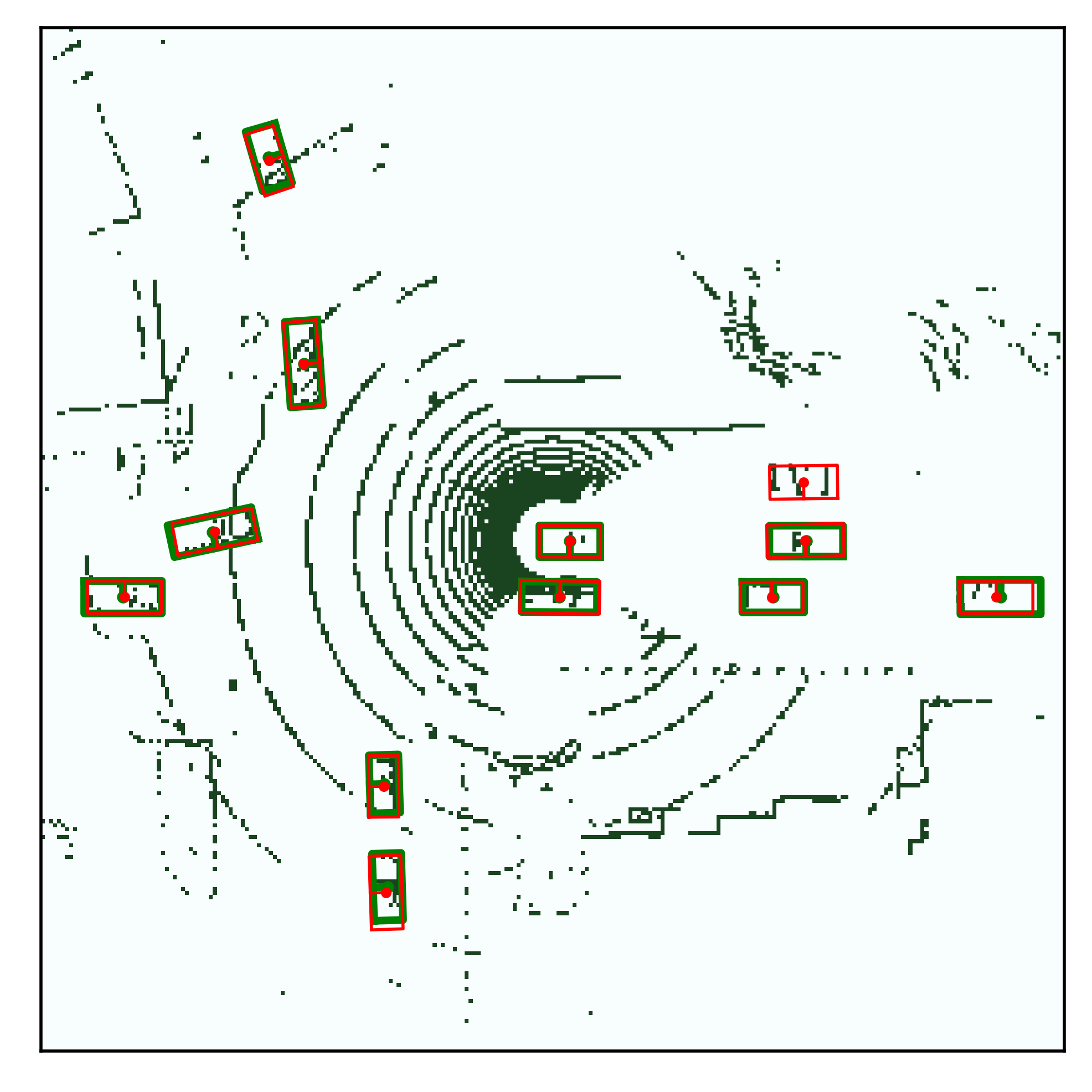}
    \includegraphics[width=\textwidth]{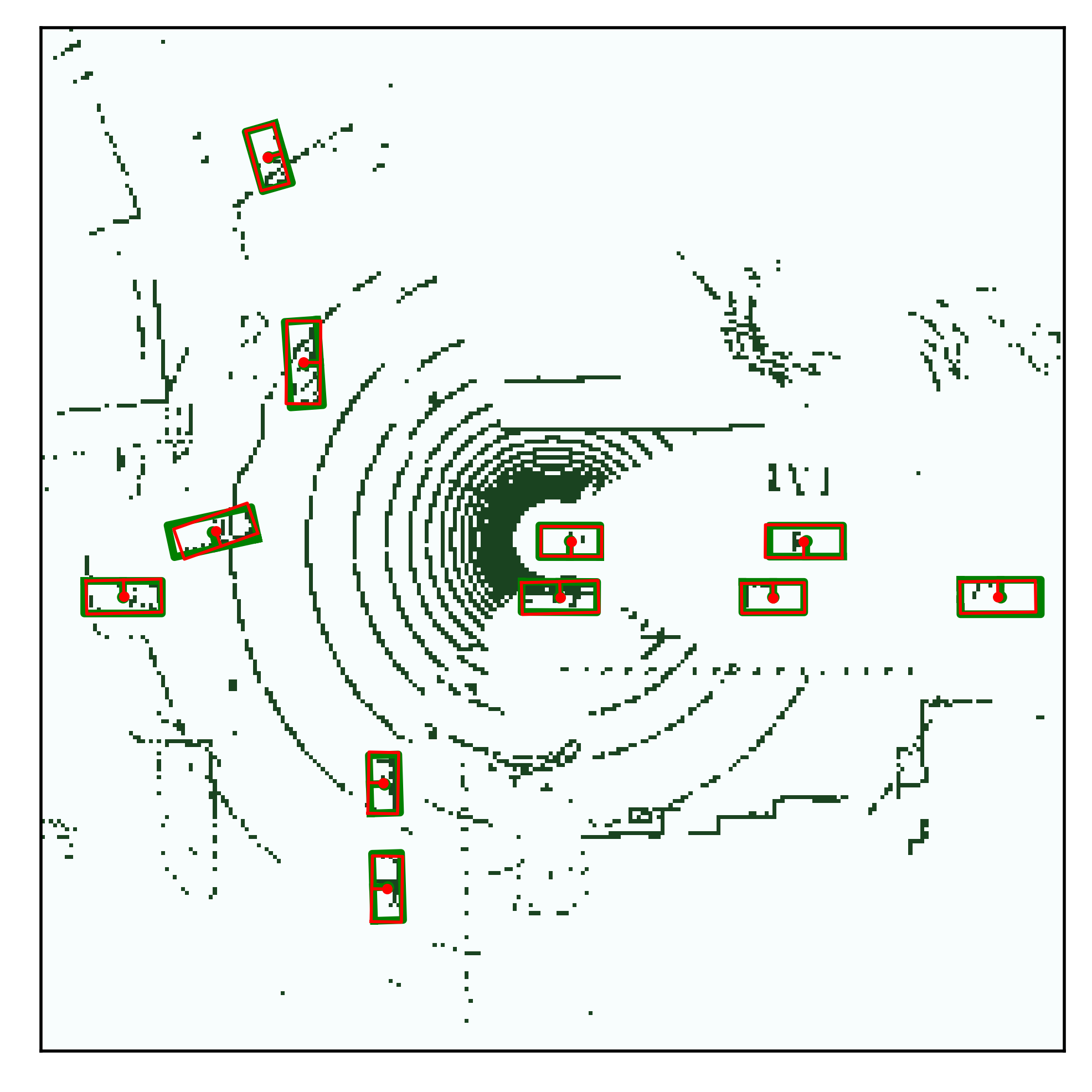}
    \caption{Scene 91 frame 53.}
    \label{fig:scene91_frame53} 
\end{subfigure}
\begin{subfigure}[t]{0.24\textwidth}
    \includegraphics[width=\textwidth]{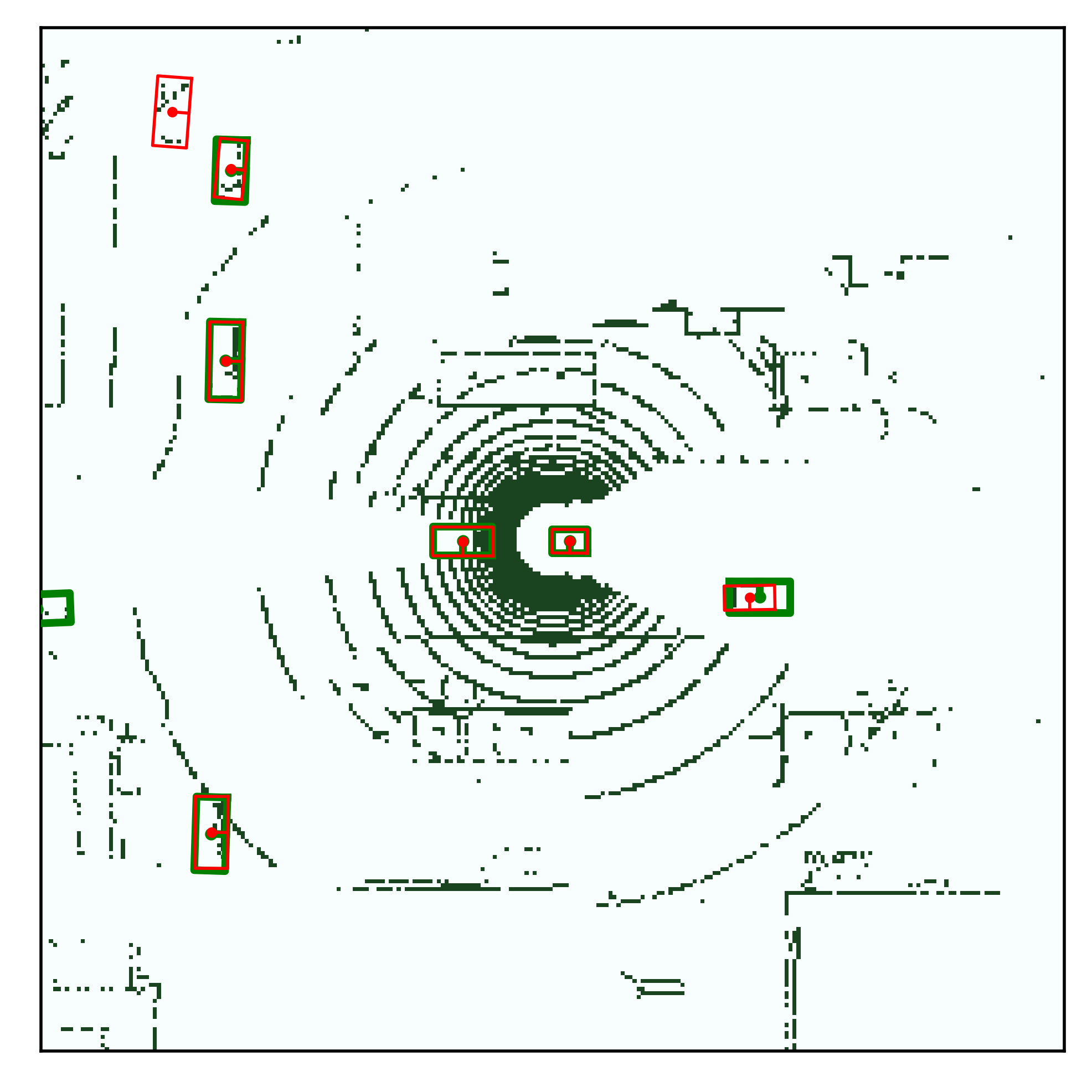}
    \includegraphics[width=\textwidth]{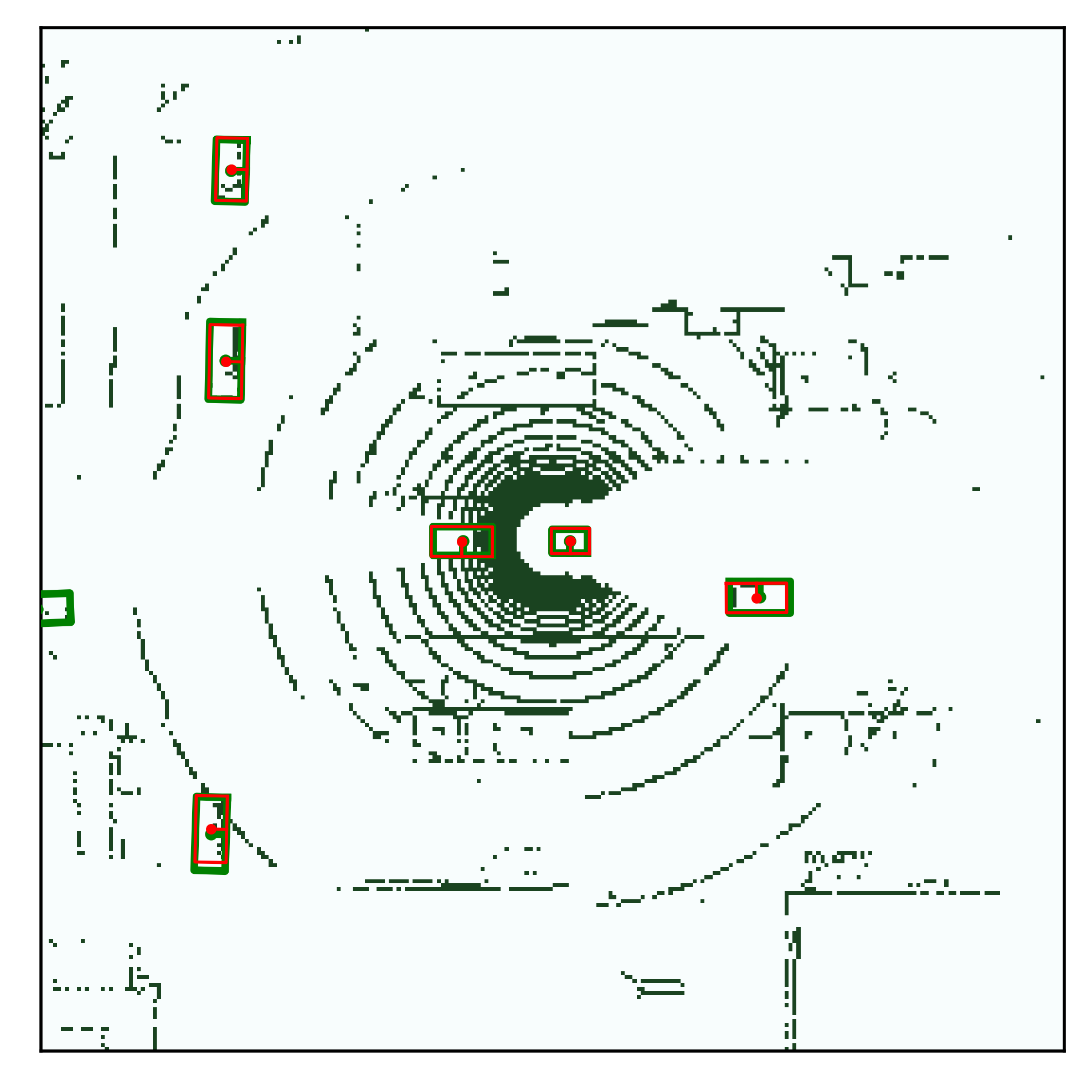}
    \caption{Scene 92 frame 74.}
    \label{fig:scene92_frame74} 
\end{subfigure}
\caption{Visualization of the object detection results of base model  (\texttt{DiscoNet (PGD Test)}) and \texttt{DiscoNet} enhanced with our \name{} (\texttt{DiscoNet+\name{}}), both under the same adversarial attack in the testing stage. Red boxes are predictions, and green boxes are ground truth.}
  \label{fig:disconet_pgd_in_test_vs_disconet_cuqcp}
  \vspace{-2pt}
\end{figure*}

For a fair comparison, we apply PGD on the above base models and evaluate their performance with and without \name{}. As shown in Table~\ref{tab:main_results}, incorporating \name{} improves object detection accuracy by an average of 80.41\%. We also visualize the object detection results of the base model \texttt{DiscoNet} (\texttt{DiscoNet} (PGD Test)) and \texttt{DiscoNet} enhanced with \name{} (\texttt{DiscoNet+\name{}}), both subjected to the same adversarial attacks during the testing stage. We randomly select scene 5 frame 4, scene 8 frame 35, scene 91 frame 53, and scene 92 frame 74 for comparison. Red boxes are predictions, and green boxes are ground truth. As shown in Fig.~\ref{fig:disconet_pgd_in_test_vs_disconet_cuqcp}, \name{} can significantly improve the gap between the ground truth bounding boxes and the predicted bounding boxes from the base model under attack. Moreover, as shown in Fig.~\ref{fig:scene8_frame35}, \name{} effectively mitigates the occurrence of incorrect bounding box predictions caused by adversarial perturbations. The results indicate that, as intended, \name{} improves the resilience and objection accuracy of existing COD models and single-agent object detection models under attack. 

Compared to the original design of the base models, \name{} incorporates both adversarial training and UQ techniques to enhance model resilience against adversarial attacks. Additionally, \name{} introduces an uncertainty loss term into the base model’s original loss function to guide the learning process, thereby reducing estimated uncertainty and improving object detection accuracy. 

\begin{table}[]
\centering
\caption{Prediction results and performance comparison on testing dataset when with and without our \textbf{\name}. Our \name{} demonstrates a 80.41\% improvement in object detection accuracy compared to the base models under attacks.}
\renewcommand{\arraystretch}{1.2}
\vskip -0.1in
\label{tab:main_results}
\resizebox{\linewidth}{!}{\begin{tabular}{|l||c|c|c|c|}
\hline
Scheme  &  $\texttt{AP@IoU=0.5} \uparrow$ & $\texttt{AP@IoU=0.7} \uparrow$ &  $\texttt{KLD} \downarrow$ & $\texttt{NLL} \downarrow$  \\ \hline
\texttt{V2VNet} (PGD Test) & 28.23 & 24.23 & - & - \\ \hline
\texttt{V2VNet+\name{}} & \textbf{51.65} & \textbf{44.04} & 315.05 & 315.97 \\ \hline \hline

\texttt{DiscoNet} (PGD Test) & 25.27 & 23.56 & - & - \\ \hline
\texttt{DiscoNet+\name{}} & \textbf{47.74} & \textbf{42.08} & 466.73 & 467.65 \\ \hline \hline

\texttt{When2com} (PGD Test) & 17.57 & 16.44 & - & - \\ \hline
\texttt{When2com+\name{}} & \textbf{29.50} & \textbf{27.85} & 189.56 & 190.48 \\ \hline \hline

\texttt{CoMamba} (PGD Test) & 29.32 & 24.64 & - & - \\ \hline
\texttt{CoMamba+\name{}} & \textbf{52.51} & \textbf{45.32} & 310.32 & 310.20 \\ \hline \hline

\texttt{Upper-bound} (PGD Test) & 30.09 & 26.75 & - & - \\ \hline
\texttt{Upper-bound+\name{}} & \textbf{53.61} & \textbf{51.72} & 288.81 & 290.12 \\ \hline \hline

\texttt{Lower-bound} (PGD Test) & 22.45 & 19.83 & - & - \\ \hline
\texttt{Lower-bound+\name{}} & \textbf{31.46} & \textbf{29.21} & 250.32 & 251.51 \\ \hline

\end{tabular}}
\end{table}


\subsection{Ablation Studies} \label{subsec:ablation_studies}

$\bullet$ \textbf{Different module combinations against adversarial attacks.}
To assess the effect of each component of \name{}, we conduct experiments using the following configurations:
\begin{itemize}
   \item Default \texttt{DiscoNet} (no attack)
   \item Default \texttt{DiscoNet} under PGD attack  (\texttt{DiscoNet} (PGD Test))
   \item Integrating the learning-based UQ component to \texttt{DiscoNet}: \texttt{DiscoNet+DM}
   \item \texttt{DiscoNet+DM} under PGD attack during the testing stage: \texttt{DiscoNet+DM} (PGD Test)
   \item Integrating the learning-based UQ component and adversarial training (PGD) to \texttt{DiscoNet}, and applying PGD attack during the test stage: \texttt{DiscoNet+DM} (PGD Train+Test)
   \item Integrating our \name{} to \texttt{DiscoNet}, but no PGD attack during the testing stage: \texttt{DiscoNet+TUQCP} (PGD train)
   \item \texttt{DiscoNet+TUQCP}, whose default configuration incorporates PGD in both the training and testing stages.
\end{itemize}

As shown in Table~\ref{tab:ablation_study_dn}, integrating the learning-based UQ component with the base model enhances object detection accuracy by 63.84\% compared to the base model when subject to the same adversarial attack (\texttt{DiscoNet+DM (PGD Test)} vs \texttt{DiscoNet} (PGD Test)). Combining adversarial training and the learning-based UQ component to the base model increases object detection accuracy by 7.92\% and reduces object detection uncertainty by 38.59\%, compared to using the learning-based UQ component alone under the same adversarial attack (\texttt{DiscoNet+DM} (PGD Train+Test) vs \texttt{DiscoNet+DM (PGD Test)}). Furthermore, calibrating the estimated uncertainty by conformal prediction improves object detection accuracy by 3.9\% and reduces detection uncertainty by 5.20\%, compared to \name{} without conformal prediction under the same adversarial attack (\texttt{DiscoNet +\name{} } vs \texttt{DiscoNet+DM (PGD Train+Test)}). Most importantly, all component combinations of \name{}, particularly the learning-based UQ component, significantly enhance object detection accuracy and reduce uncertainty compared to the base model under the same adversarial attack. These results demonstrate the resilience and effectiveness of \name{} against adversarial attacks.

$\bullet$ \textbf{Model robustness against different adversarial attacks.} While existing adversarial training methods enhance model resilience against adversarial attacks, their effectiveness is typically limited to the specific types of attacks encountered during the training phase. However, real-world adversarial attacks often differ from those used during training, rendering object detection models less effective in such scenarios. To solve this, our proposed \name{} models the collaborative object detection uncertainty of the base model under different adversarial attacks. To validate this, we evaluate the object detection performance of trained \texttt{DiscoNet+\name{}} against different PGD objective functions, namely (i) PGD using classification loss $\mathcal{L}_{cls}$ as shown in Algorithm~\ref{alg:PGD}, (ii) PGD using regression loss $\mathcal{L}_{reg}$, (iii) PGD loss using localization and classification losses~\cite{im2022adversarial}. 
As shown in Table~\ref{tab:different_adversarial_attacks}, \name{} achieves a comparable object detection accuracy and uncertainty estimation when tested against various adversarial attacks during the testing stage. This demonstrates the robustness of \name{} against different adversarial attacks.

\begin{table}[]
\centering
\caption{Object detection performance of \texttt{DiscoNet+\name{}} against different PGD objective functions.}
\renewcommand{\arraystretch}{1.2}
\vskip -0.1in
\label{tab:different_adversarial_attacks}
\resizebox{\linewidth}{!}{\begin{tabular}{|l||c|c|c|c|}
\hline
Scheme  &  $\texttt{AP@IoU=0.5} \uparrow$ & $\texttt{AP@IoU=0.7} \uparrow$&  $\texttt{KLD} \downarrow$ & $\texttt{NLL} \downarrow$ \\ \hline
\texttt{DiscoNet+\name{} (PGD (Cla) Test)} & 47.74 & 42.08 & 466.73 & 467.65 \\ \hline 

\texttt{DiscoNet+\name{} (PGD (Reg) Test)} & 54.08 & 49.25 & 355.79 & 354.87 \\ \hline 

\texttt{DiscoNet+\name{} (PGD (Cla+Loc) Test)}  & 47.74 & 42.08 & 462.80 & 463.72 \\ \hline
\end{tabular}}
\vspace{-0.1in}
\end{table}

\subsection{Sensitivity Studies} \label{subsec:sensitivity_studies}

$\bullet$ \textbf{Effects of varying PGD parameter values on object detection performance.} Different settings in the adversarial training can significantly impact the object detection performance. In PGD, increasing the number of adversarial iterations $K$ can result in a more refined and effective attack, while reducing $K$ weakens the attack. A larger $\epsilon$ allows for stronger perturbations. To figure out the varying impacts of PGD settings, we set the adversarial training learning rate $\eta$ as 0.1, perturbation budget $\epsilon$ as $\{0.1, 0.5, 0.9\}$, adversarial training iteration $K$ values as $\{15, 25, 35\}$, respectively. As shown in Table~\ref{tab:pgd_eval}, the performance of our \name{} decreases when the attackers generate higher quality attacks. However, \name{} still significantly outperforms the default \texttt{DiscoNet} under adversarial attacks, as evidenced in Table~\ref{tab:ablation_study_dn}. Additionally, our \name{} performs better when the perturbations remain moderately noticeable ($\epsilon$ increasing from 0.1 to 0.5). However, when subjected to an extremely strong adversarial attack ($\epsilon=0.9$), the model struggles to distinguish malicious information, leading to further degradation in detection performance.
\begin{table}[]
\centering
\caption{Prediction results and performance comparison on utilizing different component combinations of \name{}.}
\renewcommand{\arraystretch}{1.2}
\vskip -0.1in
\label{tab:ablation_study_dn}
\resizebox{\linewidth}{!}{\begin{tabular}{|l||c|c|c|c|}
\hline
Scheme  &  $\texttt{AP@IoU=0.5} \uparrow$ & $\texttt{AP@IoU=0.7} \uparrow$&  $\texttt{KLD} \downarrow$ & $\texttt{NLL} \downarrow$ \\ \hline
\texttt{DiscoNet} & 70.84 & 64.43 & - & -  \\ \hline
\texttt{DiscoNet(PGD Test)} & 25.27 & 23.56 & - & - \\ \hline
\texttt{DiscoNet+DM} & 68.82 & 63.79 & 620.25 & 619.33 \\ \hline
\texttt{DiscoNet+DM(PGD Test)} & 40.94 & 39.03 & 802.04 & 802.96 \\ \hline 
\texttt{DiscoNet+DM(PGD Train+Test)} & 45.12 & 41.23 & 491.43 & 494.21 \\ \hline


\texttt{DiscoNet+\name{}(PGD Train)} & 54.48 & 46.61 & 417.20 & 418.12 \\ \hline
\texttt{DiscoNet+\name{}} & 47.74 & 42.08 & 466.73 & 467.65 \\ \hline
\end{tabular}}
\vspace{-0.1in}
\end{table}

\begin{table*}
\centering
\caption{Impacts of different PGD settings on \texttt{DiscoNet+\name{}}.}
\renewcommand{\arraystretch}{1.2}
\vskip -0.1in
\label{tab:pgd_eval}
\resizebox{\linewidth}{!}{\begin{tabular}{|c|c|c||c|c|c|c||c|c|c|c|}
\hline

\multicolumn{3}{|c||}{\textbf{Settings of PGD}} & \multicolumn{4}{c||}{\textbf{\texttt{DiscoNet+\name{} (PGD Train+Test)}}} & \multicolumn{4}{c|}{\textbf{\texttt{DiscoNet+\name{}(PGD Train)}}} \\ \hline

$\eta$ & $\epsilon$ & $K$ & $\texttt{AP@IoU=0.5} \uparrow$ & $\texttt{AP@IoU=0.7} \uparrow$ &  $\texttt{KLD} \downarrow$ & $\texttt{NLL} \downarrow$ &  $\texttt{AP@IoU=0.5} \uparrow$ & $\texttt{AP@IoU=0.7} \uparrow$ &  $\texttt{KLD} \downarrow$ & $\texttt{NLL} \downarrow$\\ \hline

0.1 & 0.1 & 25  & 41.39 & 33.59 & 586.89 & 587.80 
 & 41.43 & 33.87 & 567.43 & 568.35 \\ \hline
0.1 & 0.5 & 25 & 47.74 & 42.08 & 466.73 & 467.65 & 54.48 & 46.61 & 417.20 & 418.12 \\ \hline
0.1 & 0.9 & 25 & 17.70 & 13.99 & 871.12 & 872.04 & 17.71 & 14.42 & 874.23 & 875.15 \\ \hline \hline

0.1 & 0.5 & 15 & 54.04 & 49.87 & 132.78 & 133.70 & 56.25 & 52.42 & 129.65 & 130.57 \\ \hline
0.1 & 0.5 & 35 & 41.301 & 34.82 & 578.73 & 579.65 & 42.87 & 36.71 & 580.55 & 579.63 \\ \hline 

\end{tabular}}
\vspace{-0.1in}
\end{table*}

\section{Conclusion \& Discussion}\label{sec:conclusion_discussion} 

In this work, we proposed the \name{} framework to enhance the resilience of Collaborative Object Detection (COD) model under adversarial attacks. We highlighted the efficacy of Uncertainty Quantification (UQ) techniques in COD against PGD attacks. More specifically, we proposed a learning-based UQ component in conjunction with adversarial training. When integrated to COD models, \name{} improved not only the object detection accuracy but also better object detection uncertainty estimation. 
We demonstrated the benefits of \name{} using the use case of automated driving. In further work, we will investigate its applicability to other domains such as drones or real-time 3D rendering.



\newpage

\bibliographystyle{named}
\bibliography{main}



\end{document}